\documentclass[runningheads]{llncs}
\usepackage{graphicx}
\usepackage{subfig}
\usepackage{adjustbox}
\usepackage{booktabs}
\usepackage{hyperref}
\usepackage{url}
\usepackage{xcolor}
\usepackage{multirow}
\usepackage{booktabs}
\usepackage{siunitx}
\usepackage{enumerate}
\usepackage{enumitem}
\usepackage{amsmath}
\usepackage{color}
\usepackage{lineno}
\usepackage{slashbox}
\usepackage{soul}

\begin{document}
\title{Classifying Math Knowledge Components via Task-Adaptive Pre-Trained BERT}

%
\author{}
\institute{}


\author{Jia Tracy Shen\inst{1} \and
Michiharu Yamashita\inst{1}\and
Ethan Prihar\inst{2}\and
Neil Heffernan\inst{2} \and
Xintao Wu\inst{3}\and
Sean McGrew\inst{4} \and
Dongwon Lee \inst{1} }
\authorrunning{J. Shen et al.}
\titlerunning{Classifying Math KCs via Task-Adaptive Pre-Trained BERT}
%

\institute{Penn State University, University Park, PA 16802, USA\\
\email{jqs5443@psu.edu}\and
Worcester Polytechnic Institute, Worcester, MA 01609, USA\\
\email{ebprihar@gmail.com}\and
University of Arkansas, Fayetteville, AR 72701, USA\\
\email{xintaowu@uark.edu}\and
K12.com, Herndon, VA 20170, USA\\
\email{smcgrew@k12.com}}
\maketitle              

\begin{abstract}

Educational content labeled with proper knowledge components (KCs) are particularly useful to teachers or content organizers.
However, manually labeling educational content is labor intensive and error-prone. To address this challenge, prior research proposed machine learning based solutions to auto-label educational content with limited success.
In this work, we significantly improve prior research by (1) expanding the input types to include KC descriptions, instructional video titles, and problem descriptions (i.e., three types of prediction task), (2) doubling the granularity of the prediction from 198 to 385 KC labels (i.e., more practical setting but much harder multinomial classification problem), (3) improving the prediction accuracies by 0.5-2.3\% using Task-adaptive Pre-trained BERT, outperforming six baselines, and (4) proposing a simple evaluation measure by which we can recover 56-73\% of mispredicted KC labels. All codes and data sets in the experiments are available at: \texttt{https://github.com/tbs17/TAPT-BERT}

\end{abstract}

\section{Introduction}\label{intro}
In the math education community, teachers, Intelligent Tutoring Systems (ITSs) and Learning Management Systems (LMSs) have long focused on bringing learners to the target mastery over a set of skills, also known as {\bf Knowledge Components (KCs)}. Common Core State Standards (CCSS)\footnote{www.corestandards.org} is one of the most common categorizations of knowledge components skills in mathematics from kindergarten to high school in the United States with a full set of 385 KCs. For example, in the CCSS code \textit{7.NS.A.1}, \textit{7} stands for 7-th grade, \textit{NS} stands for the domain \textit{Number system}, \textit{A.1} stands for the standard number of the code \cite{corestandards.orgCodingCCSS}. In the process of using KCs, the aforementioned stakeholders often encounter the challenges in three scenarios: (1) teachers need to know what KCs a student is unable to master by describing the code content ($S_{1}$), (2) ITSs need to tag instructional videos with KCs for better content management ($S_{2}$), and (3) LMSs need to know what KCs a problem is associated with in recommending instructional videos to aid problem solving ($S_{3}$). 

The solutions to these scenarios typically framed the problem as the \textit{multinominal classification}--i.e., given the input text, predicts one most relevant KC label out of many KCs: $I (nput) \mapsto text$ and $O (utput) \mapsto KC$. Prior research solutions included SVM-based \cite{Karlovcec2012KnowledgeSystemb}, Non-negative Matrix Factorization (NMF) \cite{Desmarais2012MappingFactorization}, Skip-gram Representation \cite{Pardos2017ImputingContext}, Neural Network \cite{Patikorn2019GeneralizabilityTexts} or even cognitively-based knowledge representation \cite{Rose2005AutomaticAssessment}. Existing solutions, however, used relatively small number of labels (e.g., 39 or 198) from CCSS with the input of problem text only (similar to Table~\ref{data_intro}-Row 3) \cite{Pardos2017ImputingContext,Karlovcec2012KnowledgeSystemb,Patikorn2019GeneralizabilityTexts}.

\begin{table}[tb]
\caption{Examples of three data types, all having the KC label ``8.EE.A.1" }
\label{data_intro}
\centering
\begin{tabular}{|c|c|}
\hline
Data Type & Text\\
\hline
\multirow{2}{*}{Description Text} & Know and apply the properties of integer\\ 
&exponents to generate equivalent numerical expressions \\
\hline
\multirow{2}{*}{Video Title} & 
Apply properties of integer exponents to generate \\ &equivalent numerical expressions\\
\hline
\multirow{2}{*}{Problem Text} &Simplify the expression: (z2)2 *Put parentheses around \\
&the power if next to coefficient, for example: 3x2=3($x^2$),x5=$x^5$ \\
\hline
\end{tabular}
\vspace{-\baselineskip}
\end{table}

Toward this challenge, in this work, we significantly improve existing methods in auto-labeling educational content. First, based on three scenarios of $S_{1}$, $S_{2}$, and $S_{3}$, we consider three types of input, including KC descriptions, instructional video titles, and problem text (as shown in Table~\ref{data_intro}).
Second, we solve the multinomial classification problem with 385 KC labels (instead of 198). Note that the problem becomes much harder. Third, we adopt the \textit{Task-adpative Pre-trained} (TAPT) BERT \cite{Gururangan2020DontTasks} in solving the multinomial classification problem. Our solution outperforms six baselines, including three classical machine learning (ML) methods and two prior approaches, improving the prediction accuracies by 0.5-2.3\% for the tasks of  $S_{1}$, $S_{2}$, and $S_{3}$, respectively. Finally, we propose a new evaluation measure, \textit{TEXSTR}, that enables 56-69\% more KC labels to be correctly predicted than using the classical measure of \textit{accuracy}.

\section{Related Work}

\noindent \textbf{KC Models.} Rose et al. \cite{Rose2005AutomaticAssessment} is one of the earliest work predicting knowledge components, which took a cognitively-based knowledge representation approach.  The scale of KCs it examined was small with only 39 KCs. Later research extended the scale of KCs using a variety of techniques. For example, Desmariais \cite{Desmarais2012MappingFactorization} used non-negative matrix factorization to induce Q-matrix \cite{Birenbaum1993DiagnosingModel} from simulated data and obtained an accuracy of 75\%. The approach did not hold when applying to real data and only got an accuracy of 35\%. The two aforementioned studies shared the same drawback: not using the texts from the problems. Karlovcec et al. \cite{Karlovcec2012KnowledgeSystemb} used problem text data from the ASSISTments platform \cite{Heffernan2014TheTeaching} and created a 106-KC model using 5-fold cross validation via ML approach SVM, achieving top 1 accuracy of 62.1\% and top 5 accuracy of 84.2\%. Pardos et al. \cite{Pardos2017ImputingContext} predicted for 198 labels and achieved 90\% accuracy via Skip-gram word embeddings of problem id per user (no problem text used). However, Patikorn et al. \cite{Patikorn2019GeneralizabilityTexts} did a generalizability study of Pardos et al. \cite{Pardos2017ImputingContext}'s work and only achieved 13.67\% accuracy on a new dataset. They found that was because Pardos et al. \cite{Pardos2017ImputingContext}'s model was over-fitting due to memorizing the question templates and HTML formatting as opposed to encoding the real features of the data. Hence, Patikorn et al. \cite{Patikorn2019GeneralizabilityTexts} removed all the templates and HTML formatting and proposed a new model using Multi-Layer-Perceptron algorithm, which achieved 63.80\% testing accuracy and 22.47\% on a new dataset. The model of Patikon et al. \cite{Patikorn2019GeneralizabilityTexts} became the highest performance for the type of problem text. The preceding research is only focused on problem related content (ID or texts) whereas our work uses not only the problem text but also the KC descriptions and video title data covering a broad range of data.

\noindent \textbf{Pre-Trained BERT Models.} The state-of-the-art language model BERT (Bidirectional Encoder Representations From Transformer) \cite{Devlin2019BERT:Understanding} is a pre-trained language representation model that was trained on 16 GB of unlabeled texts including Books Corpus and Wikipedia with a total of 3.3 billion words and a vocabulary size of 30,522. Its advantage over other pre-trained language models such as ELMo \cite{Peters2018DeepRepresentations} and ULMFiT \cite{Howard2018UniversalClassificationb} is its bidirectional structure by using the \textit{masked language model} (MLM) pre-training objective. The MLM randomly masks 15\% of the tokens from the input to predict the original vocabulary id of the masked word based on its context from both directions \cite{Devlin2019BERT:Understanding}. The pre-trained model then can be used to train from new data for tasks such as text classification, next sentence prediction. 

Users can also further pre-train BERT model with their own data and then fine-tune. This combining process has become popular in the past two years as it can usually achieve better results than fine-tuning only strategy. Sun et al. \cite{Sun2019HowClassification} proposed a detailed process on how to further pre-train new texts and fine-tune for classification task, achieving a new record accuracy. Models such as FinBERT \cite{Liu2020FinBERT:Mining}, ClinicalBERT \cite{Alsentzer2019PubliclyEmbeddingsb}, BioBERT \cite{Lee2020DataMining}, SCIBERT \cite{Beltagy2019SCIBERT:Text}, and E-BERT \cite{Zhang2020E-BERT:Reconstruction} that were further pre-trained on huge domain corpora (e.g.billions of news articles, clinical texts or PMC Full-text and abstracts) were referred as \textit{Domain-adaptive Pre-trained} (DAPT) BERT and models further pre-trained on task-specific data are referred as \textit{Task-adaptive Pre-trained} (TAPT) BERT by Gururangan et al. \cite{Gururangan2020DontTasks} such as MelBERT \cite{Choi2021MelBERTTheories} (Methaphor Detection BERT). Although DAPT models usually achieve better performance (1-8\% higher), TAPT models also demonstrated competitive and sometimes even higher performance (2\% higher) according to Gururangan et al. \cite{Gururangan2020DontTasks}. In Liu et al. \cite{Liu2020FinBERT:Mining}, FinBERT-task was 0.04\% higher than domain FinBERT in accuracy. In addition, TAPT requires less time and resource to train. In light of this finding, we use the task-specific data to further pre-train the BERT model.

\section{The Proposed Approach}

To improve upon existing solutions to the problem of auto-labeling educational content, we propose to exploit recent advancements by BERT language models.
Since BERT can encode both linguistic structures and semantic contexts in texts well, we hypothesize its effectiveness in solving the KC labeling problem. By effectively labeling the KCs, we expect to  solve the challenges incurred from three scenarios in Section \ref{intro}. 

\begin{figure*}[t]
\vspace*{-\baselineskip}
\centering
 \includegraphics[width=0.8\textwidth]{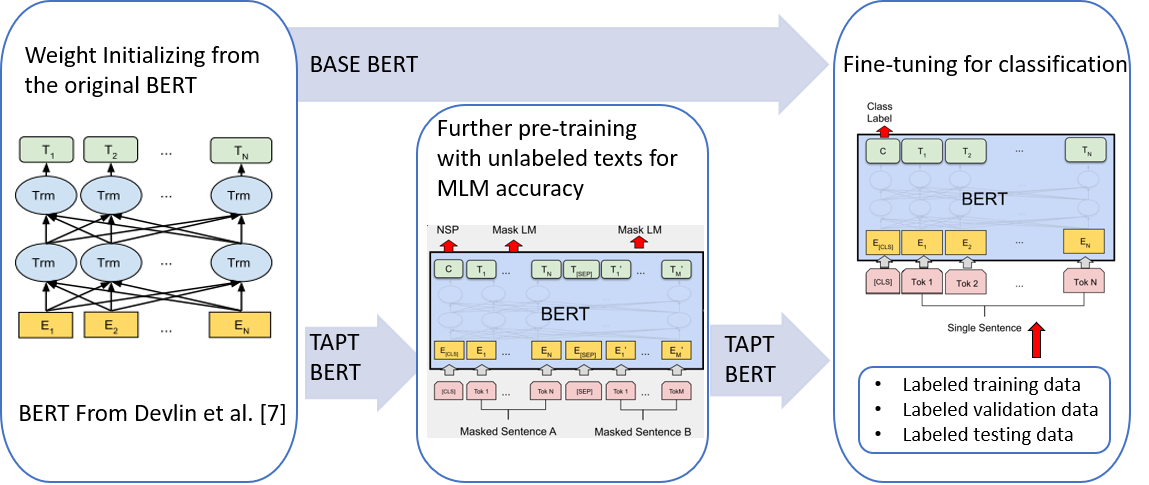}\\
\caption{An illustration of training and fine-tuning process of BASE vs. TAPT} 

\label{fu_ft}
\vspace*{-\baselineskip}
\end{figure*}
\vspace*{-\baselineskip}

\subsection{Task-Adpative Pre-Trained (TAPT) BERT}\label{train_strat}

In particular, we propose to adopt the Task-adaptive Pre-trained (TAPT) BERT and  fine-tune it for three types of data. The ``pre-training" process is  unsupervised such that unlabeled task-specific texts get trained for MLM objective whereas the ``fine-tuning" process is supervised such that labeled task-specific texts get trained for classification (see Fig. \ref{fu_ft}). We call a BERT model that only has a fine-tuning process as BASE. For TAPT, we first initialize the weights from the original BERT (i.e., BERT-base-uncased model). Then, we further pre-train the weights using the unlabeled task-specific texts as well as the combined task texts (see detail in Section \ref{data_detail}) for MLM objective, a process of randomly masking off 15\% of the tokens and predict their original vocabulary IDs. The pre-training performance is measured by the accuracy of MLM. Once TAPT is trained, we fine-tune TAPT with the task-specific labeled texts by splitting them into training, validation and testing datasets and feed them into the last softmax layer for classification. We measure the performance of fine-tuning via the testing data accuracy. For BASE, we do not further train it after initializing the weights but directly fine-tune it with the task-specific data for classification (see Fig. \ref{fu_ft}). To show the effectiveness of the TAPT BERT approach, we compare it against six baselines including BASE BERT for three tasks: 
\begin{itemize}
    \item $T_{d}$: to predict K-12 KCs using dataset $D_{d}$ (description text) based on $S_{1}$
    \item $T_{t}$: to predict K-12 KCs using dataset $D_{t}$ (video title text) based on $S_{2}$
    \item $T_{p}$:  to predict K-12 KCs using dataset $D_{p}$ (problem text) based on $S_{3}$

\end{itemize}

\subsection{Evaluating KC Labeling Problem Better: {\em TEXSTR}}\label{eval_metric}

\begin{figure*}[t]
\vspace*{-2\baselineskip}
\centering
 \includegraphics[width=0.9\textwidth]{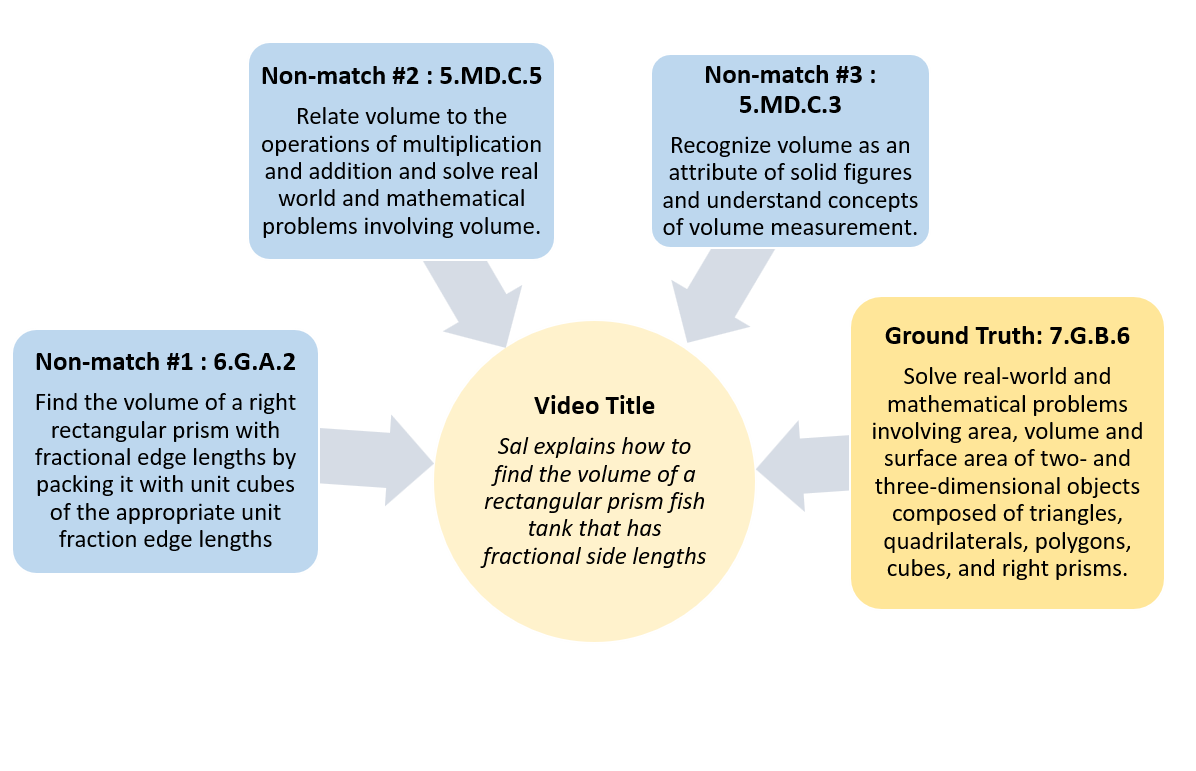}
\vspace*{-2\baselineskip}
\caption{An illustration of multiple possibilities of a correct label for a given video title text} 
\label{metric_example}
\vspace*{-\baselineskip}
\end{figure*}

In the regular setting of multinomial classification to predict KC labels, the evaluation is done as binary--i.e., exact-match or non-match. For instance, if a method predicts a KC label to be {\em 7.G.B.6}, but its ground truth is {\em 7.G.A.5}, {\em 7.G.B.6} is considered to be a non-match. However, the incorrectly predicted label of {\em 7.G.B.6} could be closely related to {\em 7.G.A.5} and thus still be useful to teachers or content organizers. For example, in Fig. \ref{metric_example}, the input to the classification problem is a video title  ``Sal explains how to find the volume of a rectangular prism fish tank that has fractional side lengths." Its ground truth label is \textit{7.G.B.6} (7-th grade geometry KC), described as ``Solve real world problem involving ... volume ... composed of ... prisms." When one looks at three non-match labels, however, their descriptions do not seem to be so different (see in Fig. \ref{metric_example}). That is, all of the three non-match labels (\textit{6.G.A.2}, \textit{5.MD.C.5}, and \textit{5.MD.C.3}) mention ``volume solving" through ``fine/relate/recognize with operations and concepts," which is quite similar to the KC description of the ground truth. However, due to the nature of exact-match based evaluation, these three labels are considered wrong predictions. Further, domain experts explain that some skills are prerequisites to other skills, or that some problems have more than one applicable skills (thus multiple labels) and they could all be correct. 

Therefore, we argue that using a strict exact-matching based method in evaluating the quality of the predicted KC labels might be insufficient in practical settings. We then propose a method that considers both semantic and structural similarities among KC labels and their descriptions to be an additional measure to evaluate the usability of the predicted labels.

\begin{itemize}
    \item Semantic Similarity ($C_t$): We adopt the Doc2Vec algorithm \cite{Le2014DistributedDocuments} to capture the similarity between KC labels. Doc2Vec, derived from word-vector algorithm, generates similarity scores between documents instead of words and is proved to have lower error rate (7.7-16\%) than the word vector approach \cite{Le2014DistributedDocuments}. 
    \item Structural Similarity ($C_s$): We exploit prerequisite relationships among skills (KC labels) and capture such as edges and KC labels as nodes in a graph.
    The prerequisite relationships are extracted from a K-G8 math coherence map by Jason Zimba \cite{Zimba2012AStandards} and a high school (G9-G12) coherence map by UnboundEd Standard Institue \cite{UnboundEd2017AStandards}. Then, we adopt Node2Vec algorithm \cite{Grover2016Node2vec:Networks} that is efficient and flexible in exploring nodes similarity and achieved a new record performance in network classification problem \cite{Grover2016Node2vec:Networks}.
\end{itemize} 

In the end, we craft a new evaluation measure, named as \textit{TEXSTR} ($\Lambda$), by combining both $C_{t}$ and $C_{s}$ as follows: $\Lambda=\alpha\cdot C_{t}+(1-\alpha)\cdot C_{s}$, where $\alpha$ controls the weight between $C_t$ and $C_s$ as an oscillating parameter.

\section{Empirical Validation} \label{emp_valid}

\subsection{Datasets and Evaluation Measure} \label{data_detail}

Table \ref{dataset} summarizes the details of the datasets  for pre-training and fine-tuning processes. $D_{d}$ contains 6,384 description texts (84,017 tokens) and 385 math KCs (an example shown in Fig. \ref{data_intro}-a). Part of $D_{d}$ are extracted from Common Core Standards website\footnote{http://www.corestandards.org/math} and part are provided by k12.com\footnote{http://www.k12.com}, an education management organization that provides online education to American students from kindergarten to Grade 12. $D_{t}$ contains 6,748 video title texts (62,135 tokens) and 272 math KCs (an example shown in Fig. \ref{data_intro}-b) Part of $D_{t}$ are extracted from  \textit{Youtube.com} (via youtube DataAPI\footnote{http://developers.google.com/youtube/v3}) and part are provided by k12.com. $D_{p}$ contains 13,722 texts (589,549 tokens) and 213 math KCs provided by ASSISTments\footnote{http://www.assistments.org/} (an example shown in Fig. \ref{data_intro}-c). Further, $D_{d+t}$, $D_{d+p}$, $D_{t+p}$, and $D_{all}$ are different combinations of the unlabeled texts from $D_{d}$, $D_{t}$, and $D_{p}$. They are only used in the TAPT pre-training process. We pre-process all aforementioned texts by removing all the templates and HTML markups to avoid over-fitting, suggested by the prior highest accuracy method \cite{Patikorn2019GeneralizabilityTexts}. In the TAPT pre-training process, 100\% of the unlabeled texts from the aforementioned datasets are used for pre-training. In fine-tuning process for both TAPT and BASE , only $D_{d}$, $D_{t}$, and $D_{p}$ are used and 72\% of their texts and labels are used for training, 8\% are for validation and 20\% are for testing (see in Table \ref{dataset} Row 1-3 and Col. 6-8). 

As an evaluation measure, following prior research \cite{Patikorn2019GeneralizabilityTexts,Pardos2017ImputingContext,Rose2005AutomaticAssessment,Desmarais2012MappingFactorization,Karlovcec2012KnowledgeSystemb} for direct comparison, we use Accuracy@k as (TP + TN)/(TP + TN + FP + FN), when a method predicts top-$k$ KC labels. Further, we evaluate our method using the proposed {\em TEXSTR} measure.
\begin{table}[tb]
\caption{A summary statistics of datasets. 
}\label{dataset}
\centering
\begin{tabular}{|c|c|c|c|c|c|c|}
\hline
\multirow{2}{*}{Name} &\multirow{2}{*}{\# Labels}&\multirow{2}{*}{\# Texts}&
\multirow{2}{*}{\#  Tokens} &
\multicolumn{3}{c|}{Fine-tuning Partition}\\ \cline{5-7}
&&&&Training (72\%)&Validation (8\%)&Testing (20\%)\\
\hline
$D_{d}$ &385 & 6,384&84,017&4,596&511&1,277 \\
$D_{t}$&272&6,748 &62,135&4,858&540&1,350\\
$D_{p}$&213&13,722&589,549&9,879&1,098&2,745\\
\hline
$D_{d+t}$&/ &13,132&146,152&/&/&/\\
$D_{d+p}$ &/&20,106&673,566&/&/&/\\
$D_{t+p}$ &/ &20,470&651,684&/&/&/\\

$D_{all}$&/ &26,854&735,701&/&/&/\\
\hline
\end{tabular}
\vspace*{-1.1\baselineskip}
\end{table}

\subsection{Pre-training and Fine-tuning Details}

To further pre-train, we follow the same pre-training process of original BERT with the same network architecture (12 layers, 768 hidden dimensions, 12 heads, 110M parameters) but on our own unlabeled task-specific texts (see Col. 4 in Table \ref{dataset}). With an 8-core v3 TPU, we further train all our models with 100k steps, achieving MLM accuracy of above 97\% that lasts about 1-4 hours. We experiment hyper-parameters such as learning rate (lr) $\in {\{1e-5, 2e-5, 4e-5, 5e-5, 2e-4\}}$, batch size (bs) $\in {\{8, 16,32\}}$, and max-sequence length (max-seq-len) $\in {\{128, 256, 512\}}$. The highest MLM accuracy was achieved when lr $\leftarrow$ 2e-5, bs $\leftarrow$ 32, and max-seq-len $\leftarrow$ 128 (for $D_{d}$ and $D_{t}$) and max-seq-len $\leftarrow$ 512 with the same lr and bs (for $D_{p}$, $D_{d+p}$, $D_{t+p}$, $D_{all}$).
To fine-tune, we also follow the original BERT script by splitting $D_{d}$, $D_{t}$, $D_{p}$ into 72\% for training, 8\% for validation and 20\% for testing per task. We experiment ep $\in {\{5, 10, 25\}}$ due to the small size of the data size and retain the same hyper-parameter search for lr, bs, max-seq-len. We find that the best testing accuracy is obtained when ep $\leftarrow$ 25, lr $\leftarrow$ 2e-5, bs $\leftarrow$ 32, and max-seq-len $\leftarrow$ 128 for $D_{d}$, $D_{t}$ whereas the best testing accuracy for $D_{p}$ is obtained when ep $\leftarrow$ 25, lr $\leftarrow$ 2e-5, bs $\leftarrow$ 32, and max-seq-len $\leftarrow$ 512. We find that after ep $\leftarrow$ 25, it is difficult to gain significant increase on the testing accuracy. Hence, the optimal hyper-parameters while task-dependent seem to have very minimal change across tasks. This finding is consistent with SCIBERT reported \cite{Beltagy2019SCIBERT:Text}.

\subsection{Result \#1: TAPT BERT vs. Other Approaches}
Table \ref{fitv.fut} summarizes the experimental results of six baseline approaches and TAPT for each task. For baseline methods, we group them into categories (see in Table \ref{fitv.fut}) (1) classical ML, (2) prior work, and (3) BASE BERT. By including popular ML methods such as Random Forest and XGBoost, we aim to compare its performance to the one from prior ML work (SVM) proposed by Karlovec et al \cite{Karlovcec2012KnowledgeSystemb} in the literature review. As to comparing to the prior highest accuracy method \cite{Patikorn2019GeneralizabilityTexts}, we applied the same 5-fold cross-validation on our own problem texts and obtain Acu@1 and Acu@3.
Overall, we see that TAPT models outperform all other methods at both Acu@1 and Acu@3 across three tasks. Note TAPT models here are simply trained on the unlabeled texts from $D_{d}$, $D_{t}$, and $D_{p}$.
Compared to the best method in baseline, TAPT has an increase of 0.70\%, 1.72\%, 0.07\% at Acu@1  and 0.51\%, 2.28\%, 1.52\% at Acu@3 across three tasks. Compared to BASE, TAPT shows an increase of 2.30\%, 1.72\%, 0.70\% at Acu@1  and 0.51\%, 2.28\%, 1.52\% at Acu@3 across  three tasks. Acu@1 and Acu@3 from both TAPT and BASE models are the average performance over five random seeds with significant difference (see last row in Table \ref{fitv.fut}). BERT variants such as FinBERT \cite{Liu2020FinBERT:Mining}, SCIBERT \cite{BeltagSCIBERT:Text}, BioBERT \cite{Lee2020DataMining} and E-BERT \cite{Zhang2020E-BERT:Reconstruction} were able to achieve a 1-4\% increase when further trained on much larger domain knowledge corpus (i.e. 2-14 billion tokens). Our corpus although comparatively small with $D_{d}$ (84,017 tokens), $D_{t}$ (62,135 tokens), and $D_{p}$ (589,549 tokens) still result in a decent improvement of 0.51-2.30\%.

\begin{table}[tb]
\caption{
Accuracy comparison (best and 2nd best accuracy in blue bold and underlined, respectively, $BL\dag$ for baseline best, and * for statistical significance with p-value $<$ 0.001)}

\label{fitv.fut}
\centering
\begin{tabular}{|c|c|c|c|c|c|c|c|}
\hline
\multirow{2}{*}{Approach Type} &
\multirow{2}{*}{Algorithm}&
\multicolumn{2}{c|}{$D_{d}$} &
\multicolumn{2}{c|}{$D_{t}$}&
\multicolumn{2}{c|}{$D_{p}$}\\
\cline{3-8}
&&Acu@1&Acu@3&Acu@1&Acu@3&Acu@1&Acu@3\\
\hline

\multirow{3}{*}{Classical ML}
&SVM \cite{Karlovcec2012KnowledgeSystemb} & 44.87 &70.40&48.15&70.30&78.07&87.69\\
&XGBoost &43.07 &71.34&45.33 &66.15&77.63 &87.94 \\
&Random Forest &49.26 &\textcolor{blue}{\underline{78.78}}&49.33 &74.37&78.03 &88.23 \\

\hline
\multirow{2}{*}{Prior Work}

&Skip-Gram NN \cite{Pardos2017ImputingContext}  & 34.07 &34.15 &43.00&43.52&76.88&77.06\\

&Sklearn $MLP$ \cite{Patikorn2019GeneralizabilityTexts} &\textcolor{blue}{\underline{50.53}}&74.41  &48.22&57.95 &80.70&81.13\\

\hline
\multirow{2}{*}{BERT}&BASE &48.30&76.40&\textcolor{blue}{\underline{ 50.99}}&\textcolor{blue}{\underline{76.55}} &\textcolor{blue}{\underline{81.73}}&\textcolor{blue}{\underline{90.99}}\\

&TAPT &\textcolor{blue}{\textbf{50.60}}&\textcolor{blue}{\textbf{79.29}}&\textcolor{blue}{\textbf{52.71}}&\textcolor{blue}{\textbf{78.83}}&\textcolor{blue}{\textbf{82.43}}&\textcolor{blue}{\textbf{92.51}}\\

\hline

\multirow{2}{*}{Improvement}

&$|TAPT-BL\dag|$ &0.07&0.51&1.72&2.28&0.70&1.52\\
&$|TAPT-BASE|$&$2.30^{*}$&$0.51^{*}$&$1.72^{*}$&$2.28^{*}$&$0.70^{*}$&$1.52^{*}$\\
\hline
\end{tabular}
\vspace*{-\baselineskip}
\end{table}

\subsection{Result \#2: Augmented TAPT and TAPT Generalizability}
In addition to the simply trained TAPTs (referred as simple TAPT) in Table \ref{fitv.fut}, we augment the pre-training data and form another four TAPTs ($TAPT_{d+t}$, $TAPT_{d+p}$, $TAPT_{t+p}$ and $TAPT_{all}$). We call them augmented TAPT.  Table \ref{across-task} showcases the differences in Acu@3  between simple and augmented TAPT. For $D_{d}$, augmented $TAPT_{d+p}$ outperforms all  simple TAPT models (Acu@3 = 79.56\%) and augmented $TAPT_{d+t}$ achieves the second best Acu@3 (79.40\%). For $D_{t}$, all the augmented TAPT models only outperform simple $TAPT_{p}$. For $D_{p}$, augmented $TAPT_{t+p}$ outperforms all simple TAPTs with Acu@3 of 92.64\%. To sum up, augmenting the pre-training data for TAPT seems to help increase the accuracy further.

\begin{table}
\centering
\caption{Acu@3: BASE vs. TAPT. (best and 2nd best per row in bold and underlined, and subscripts indicate outperformance over BASE)
}
\label{across-task}
\begin{tabular}{|c|c|c|c|c|c|c|c|c|}
\hline
\multirow{2}{*}{Data}&
\multirow{2}{*}{BASE}&
\multicolumn{3}{c|}{Simple}&
\multicolumn{4}{c|}{Augmented}\\ \cline{3-9}
&&$TAPT_{d}$&$TAPT_{t}$&$TAPT_{p}$&$TAPT_{d+t}$&$TAPT_{d+p}$&$TAPT_{t+p}$&$TAPT_{all}$\\
\hline
$D_{d}$&76.40&$79.29_{2.89}$&$78.78_{2.38}$&$77.84_{1.44}$&$\underline{79.40}_{3.00}$&$\textbf{79.56}_{3.16}$&$79.01_{2.61}$&$79.01_{2.61}$\\
$D_{t}$&76.55&$\underline{77.85}_{1.30}$&$\textbf{78.83}_{2.28}$&$76.30_{-0.25}$&$77.56_{1.01}$&$77.56_{1.01}$&$77.70_{1.15}$&$77.78_{1.23}$\\
$D_{p}$&90.99&$91.22_{0.23}$&$91.44_{0.45}$&$\underline{92.51}_{1.52}$&$92.06_{1.07}$&$92.50_{1.51}$&$\textbf{92.64}_{1.65}$&$92.35_{1.36}$\\
\hline
\end{tabular}
\vspace*{-\baselineskip}
\end{table}

Furthermore, we compare the generalizability of TAPT to BASE over different datasets. We define the \textit{generalizability} as task accuracy (specifically Acu@3) that a model can obtain when applied to a different dataset. Both BASE and TAPT are pre-trained models and obtain task accuracy via fine-tuning on a different task data. The subscripts in Table \ref{across-task}  present the difference in Acu@3 between TAPT and BASE, showcasing who has stronger generalizability ($-$ sign indicates weak generalizability). 
For $D_{d}$, all  simple and augmented TAPT models generalize better than BASE, especially augmented TAPTs have an average of about 3\% increase.  For $D_{t}$, all TAPT models have better generalizability than BASE with over 1\% average increase except for $TAPT_{p}$. For $D_{p}$, we also see all the TAPTs generalize better than BASE model with the augmented $TAPT_{t+p}$ having the best generalizability. 

\subsection{Result \#3: TEXSTR Based Evaluation}
Following the definition of \textit{TEXSTR} (=$\Lambda$) in Section \ref{eval_metric}, we vary the values of $\alpha$ by \{0, 0.5, 1\} and generate three variations of $\Lambda$  for top-3 predictions. We then decide the percentage of miss-predictions to be reconsidered based on $\Lambda$ value by three cut-off  thresholds \{0.5, 0.75, 0.9\}. Before that, we make sure that the predicted labels are not subsequent to the ground truth, i.e., if the ground truth is \textit{7.G.A.2}, a predicted label such as \textit{8.G.A.3} shall not be reconsidered as correct because it is the skill to be learned subsequently ``after" \textit{7.G.A.2}. In such a case, we exclude predicted labels that have subsequent relations to the ground truth and calculate $\Lambda$. Table \ref{prereq} presents the percentage of miss-predictions after removing the subsequent-relation labels by three $\Lambda$ thresholds when $\alpha \in \{0, 0.5, 1\}$. Across three values of $\alpha$ and datasets, note that 56-73\% of miss-predictions could be reconsidered as correct if $\Lambda>0.5$, 5-53\% of them could be reconsidered if $\Lambda>0.75$, and 0-32\% could be reconsidered if $\Lambda>0.9$. The wide percentage range for $\Lambda \in \{0.75, 0.9\}$ infers that higher thresholds of $\Lambda$ are more sensitive to the change of $\alpha$.
 
To further ensure the \textit{TEXSTR} measure to be useful in practice, we conduct an empirical study
where eight experienced K-12 math teachers rate each pair of top-3 KC labels and the corresponding text (e.g., description, video title, or problem text) on a scale of 1 to 5.
The Fleiss' kappa value to assess the multi-rater agreement among eight teachers is 0.436, which is considered as moderate agreement by Landis et al. \cite{Landis1977TheDatab}. We ensure that none of top-3 miss-predicted KCs are subsequent to ground truths and have $\Lambda$ score at least 0.5. 
Then, we quantify the \textit{relevance} ($\Upsilon$) score as either $\Lambda$ score (when $\alpha=0.5$) or teachers' rating of [1,5] range divided by 5 (to be on the same scale as \textit{TEXSTR}'s [0,1]). Table \ref{tex_tea} summarizes  three varying relevance  scores ($\Upsilon \in \{0.5, 0.75, 0.9\}$)  on the pair of top-3 predictions and the texts.  For Top-1 predictions, \textit{TEXSTR} considers all of them to have $\Upsilon>0.5$ (due to the pre-selection) and 37.93\% of all have $\Upsilon>0.75$ and 3.45\% have $\Upsilon>0.9$. Teachers, on the other hand, think that only 54.31\% of the texts  have $\Upsilon>0.5$ ($\downarrow45.69\%$ from $\Lambda$) but 43.53\% have $\Upsilon>0.75$ ($\uparrow5.6\%$ from $\Lambda$) and 31.03\%  have $\Upsilon>0.9$ ($\uparrow27.58\%$ from $\Lambda$). We also find a similar pattern for Top-2 and Top-3 predictions where teachers find 6.47-6.89\% more cases than \textit{TEXSTR} that have $\Upsilon>0.75$ and  9.48-13.79\% more cases than \textit{TEXSTR} that have $\Upsilon>0.9$. This indicates that \textit{TEXSTR} is more conservative than teachers in judging the relevance of KC labels to texts when $\Upsilon \in \{0.75, 0.9\}$, suggesting \textit{TEXSTR} is effective in reassessing miss-predictions and ``recover" them as correct labels in practice.

\begin{table}[tb]
\caption{\% of miss-predictions recovered by \textit{TEXSTR} ($\Lambda$)
}
\centering
\label{prereq}
\begin{tabular}{|c|c|c|c|c|c|c|c|c|c|c|}
\hline
\multirow{2}{*}{Data}&
\multirow{2}{*}{\# Miss-predictions}&
\multicolumn{3}{c|}{$\Lambda>0.5$} &
\multicolumn{3}{c|}{$\Lambda>0.75$} &
\multicolumn{3}{c|}{$\Lambda>0.9$} \\ \cline{3-11}
&&$\alpha=0$&$\alpha=0.5$&$\alpha=1$&$\alpha=0$&$\alpha=0.5$&$\alpha=1$&$\alpha=0$&$\alpha=0.5$&$\alpha=1$\\
\hline
 
$D_{d}$ &248&70.16&68.95&72.98&52.82&24.19&8.87&32.26&2.42&0.81\\
$D_{t}$ &240&58.33&55.83&57.5&37.92 &17.08&6.67&17.08&0&1.25\\
$D_{p}$&166 &60.84&56.63&58.43&38.55&16.27&5.42&18.67&1.2&1.2\\

\hline
\end{tabular}
\vspace*{-\baselineskip}
\end{table}

\begin{table}[tb]
\caption{\% of top-3 predictions by relevance ($\Upsilon$) level when $\alpha=0.5$
}\label{tex_tea}
\centering
\begin{tabular}{|c|c|c|c|c|c|c|c|c|c|}
\hline 
\multirow{2}{*}{$\Upsilon$}&
\multicolumn{3}{c|}{Top 1} &
\multicolumn{3}{c|}{Top 2} &
\multicolumn{3}{c|}{Top 3} 

\\ \cline{2-10}
 & $\Lambda$ & Teachers &$\Delta$& $\Lambda$ & Teachers &$\Delta$& $\Lambda$ & Teachers&$\Delta$ \\
 \hline
$>0.5$&100&54.31&-45.69&100&40.95&-59.05&100&21.98&-78.02 \\
$>0.75$ &37.93&43.53&5.60&20.69&27.16&6.47&6.9&13.79&6.89\\
$>0.9$& 3.45&31.03&27.58&0&13.79&13.79&0&9.48&9.48\\
\hline
\end{tabular}
\vspace*{-\baselineskip}
\end{table}

\section{Conclusion}
The paper classified 385 math knowledge components from kindergarten to 12th grade using three data sources (e.g., KC descriptions, video titles, and problem texts) via the \textit{Task-adaptive Pre-trained} (TAPT) BERT model. TAPT has achieved a new record by outperforming six baselines by up to 2\% at Acu@1 and up to 2.3\% at Acu@3. We also compared TAPT to BASE and found the accuracy of TAPT increased by 0.5-2.3\% with a significant p-value. Furthermore, the paper discovered that TAPT trained on the augmented data by combining different task-specific texts had better Acu@3 than TAPT simply trained on the individual datasets. In general, TAPT has better generalizability than BASE by up to 3\% at Acu@3 across different tasks. 
Finally, the paper proposed a new evaluation measure \textit{TEXSTR} to reassess the predicted KCs by taking into account semantic and structural similarity. \textit{TEXSTR} was able to reconsider 56-73\% of miss-predictions as correct for practical use. 
\vspace*{-\baselineskip}
\section{Acknowledgement}

The work was mainly supported by 
NSF awards (1940236, 1940076, 1940093). In addition, the work of Neil Heffernan was in part supported by NSF awards (1917808, 1931523, 1917713,  1903304, 1822830, 1759229), IES (R305A170137, R305A170243, R305A180401, R305A180401), EIR
(U411B190024) and ONR (N00014-18-1-2768) and Schmidt Futures.

%
%
%
\bibliographystyle{splncs04}
\bibliography{references.bib}

\end{document}